# A Deep Learning Approach for Network-wide Dynamic Traffic Prediction during Hurricane Evacuation

Rezaur Rahman[a] and Samiul Hasan[*/✉]

[a] *Department of Civil, Environmental, and Construction Engineering*
*University of Central Florida*
*Email: rezaur.rahman@knights.ucf.edu*

[*/✉]*Corresponding Author, Department of Civil, Environmental, and Construction Engineering*
*University of Central Florida*
*Email: samiul.hasan@ucf.edu*

*Abstract*—Proactive evacuation traffic management largely depends on real-time monitoring and prediction of traffic flow at a high spatiotemporal resolution. However, evacuation traffic prediction is challenging due to the uncertainties caused by sudden changes in projected hurricane paths and consequently household evacuation behavior. Moreover, modeling spatiotemporal traffic flow patterns requires extensive data over a longer time period, whereas evacuations typically last for 2 to 5 days. In this paper, we present a novel data-driven approach for predicting evacuation traffic at a network scale. We develop a dynamic graph convolution LSTM (DGCN-LSTM) model to learn the network dynamics of hurricane evacuation. We first train the model for non-evacuation period traffic data showing that the model outperforms existing deep learning models for predicting non-evacuation period traffic with an RMSE value of 226.84. However, when we apply the model for evacuation period, the RMSE value increased to 1440.99. We overcome this issue by adopting a transfer learning approach with additional features related to evacuation traffic demand such as distance from the evacuation zone, time to landfall, and other zonal level features to control the transfer of information (network dynamics) from non-evacuation periods to evacuation periods. The final transfer learned DGCN-LSTM model performs well to predict evacuation traffic flow (RMSE=399.69). The implemented model can be applied to predict evacuation traffic over a longer forecasting horizon (6 hour). It will assist transportation agencies to activate appropriate traffic management strategies to reduce delays for evacuating traffic.

*Index Terms*— artificial intelligence, network modeling, transfer learning, traffic prediction, spatiotemporal traffic pattern, evacuation traffic management

## 1. Introduction

Hurricanes have become more intense and frequent due to climate change and other related reasons. As a result, coastal residents of the United States are becoming more vulnerable to the impact of hurricanes. To save lives and reduce suffering of people during such events, emergency management agencies need efficient and pro-active strategies to ensure timely evacuation. Evacuation traffic management has been a major concern for transportation agencies and policy makers [1]. The concern has grown further after Hurricane Irma in 2017 when about 6.5 million residents of Florida were ordered to evacuate from major cities including Key West, Miami, and Tampa. With only two major interstate highways (I-75 and I-95) available for leaving Florida, the evacuation caused significant traffic congestion and crashes on highways, thus affecting the physical and mental health of evacuees.

Efficient evacuation traffic management requires detailed evacuation plan including proactive measures to overcome unexpected events such as traffic incidents or a change of hurricane path causing unexpected demand surge. Over the past decades, several traffic management strategies [2], [3] such as emergency shoulder use, contraflow operations, traffic signal control, route guidance etc. have been proposed to alleviate congestion during evacuation. However, a proactive deployment of these measures requires a better understanding of prevailing traffic conditions and prediction of future traffic for long-term horizon (> 1 hour). Thus, one of the critical aspects of evacuation traffic management is to understand the spatial and temporal distribution of congestion and to reliably predict future traffic condition. Based on this information, traffic managers can decide where, when, and for how long a strategy should be deployed to reduce delays.

Previous studies [2], [4]–[6] have adopted mathematical modeling and simulation-based approaches to predict evacuation traffic. However, these approaches are not robust enough against sudden demand surge associated with irregular traffic flow behavior during evacuation period. Moreover, developing such models for a large-scale network is very complex relying on many assumptions (e.g., user equilibrium). Due to such complexities, it takes a substantial amount of time to produce a stable solution to estimate network wide traffic which make these models less suitable for a real-time application.

Ubiquitous use of traffic sensing technologies can help us overcome these challenges ensuring real-time traffic prediction available for transportation agencies. In this context, a data-driven network-scale model can be developed for real-time traffic prediction. For example, in Florida almost all the interstate highways are equipped with microwave radar traffic detectors owned by Florida Department of Transportation (FDOT). These detectors allow us collect high-fidelity traffic data in real-time and analyze traffic trends in response to unexpected events. Although roadway detectors have widely been deployed in major highways in the USA, a few studies have explored these detectors for network level traffic analysis and modeling for evacuation traffic prediction [7], [8].

One of the major challenges towards developing a data-driven modeling framework is that such a framework needs to deal with the high dimensionality of the data due to spatiotemporal dependency among traffic variables and capture congestion propagation in the network. Recent advancements in neural computation such as deep learning models have created an opportunity to deal with such high dimensionality in traffic data. Moreover, these models also utilize the concept of graph theory to represent the congestion propagation in a large-scale network [9]–[13]. For instance, state-of-the-art graph convolutional neural networks use network structure (i.e., node-link relationship) to capture flow propagation inside the network and the cross-correlation among traffic variables. However, existing graph convolutional approaches do not represent the actual dynamic nature of a transportation network such as how travel time changes over time and thereby influencing the changes in traffic flow patterns. Another issue with such a network-wide model is that it needs an extensive amount of data to train the model to make a reliable prediction. Thus, it is not feasible to develop such a model for special events such as hurricane evacuation, where traffic data are available only for a few days.

In this study, we develop a novel deep learning architecture to overcome these challenges for evacuation traffic prediction. First, we develop a dynamic graph convolutional LSTM (DGCN-LSTM) model with contextual understanding of congestion propagation inside the network. To overcome the limitation of data scarcity and model overfitting issue, we train the model on regular period traffic data collected from May 2017 to August 2017. Later, we transfer this model to evacuation period with additional neural network block as a controller to predict traffic during evacuation period. The controller uses evacuation travel demand features such as distance from the evacuation zone, time to landfall, and other zonal level features to control which information from the DGCN-LSTM to be included to predict traffic during evacuation period. As such, the final evacuation prediction model has the contextual understanding of the congestion propagation inside the network; at the same time, it can discard any unnecessary information such as seasonality in traffic patterns which might be out of context during an evacuation period.

This study has made several contributions for evacuation traffic prediction using large-scale traffic detector data:

• It collects and analyzes large-scale traffic detector data during Hurricane Irma's evacuation, providing insights on network wide spatiotemporal patterns of evacuation traffic.

• It identifies the potential challenges to deal with large-scale real-time traffic data for evacuation traffic analysis and prediction.

• It develops a dynamic graph convolutional neural network model to learn the congestion propagation for a large-scale network.

• It implements a transfer learning based deep learning architecture to incorporate the contextual understanding of congestion propagation inside a network for evacuation traffic prediction.

## 2. Literature Review

### 2.1 Evacuation Traffic Modeling

Evacuation traffic management can significantly benefit from how accurately we can predict traffic in real time. However, practices in evacuation traffic management mostly focus on behavioral analysis of evacuees to understand/predict evacuation decisions [2], [14]–[17]. Previous studies investigated evacuation behavior during hurricanes focusing on understanding the factors relating to evacuation decisions [18]–[22], mobilization time [17], departure time [23], [24], and destination choice [25], [26]. Although a vast number of research has been carried out on individual-level evacuation decision-making [26]–[30] to determine evacuation traffic demand, these approaches depend on survey data that are expensive and difficult to conduct as a hurricane unfolds in real-time. Although a few studies has explored evacuation traffic patterns, these are limited to understanding operational capacity loss of highways during a hurricane evacuation [7], [8], [31], [32]. However, these studies provide several key insights on challenges in evacuation traffic modeling.

One of the major challenges in modeling evacuation traffic is that traffic patterns during evacuation period significantly differ from non-evacuation period, characterized by higher demand and irregular traffic variations [33]. Evacuation traffic demand is more uncertain due to sudden changes in hurricane path [34]. To model such uncertain demand variations, we need a demand responsive model. A few studies [5], [6] have adopted mathematical modeling and simulation-based approaches to model evacuation traffic. However, such approaches rely on certain assumptions to estimate the demand for evacuation traffic which might fail to present the actual evacuation scenario. A data-driven method can offer a viable alternative solution to overcome this issue. In recent years, a few studies have explored the application of data-driven methods for traffic prediction during a major event such as hurricane evacuation [35], [36]. However, these studies are not applicable for traffic prediction for entire transportation network.

*2.1 Application of data driven methods for network level traffic modeling*

Data driven approach predicts the traffic state by finding the current traffic state from historical traffic patterns, without any assumption on user behavior [38], which make them more robust and demand responsive. Recently, data-driven approaches are gaining more attention mainly for three reasons: first, calibrating parameters required for model-driven approaches is time consuming; second, an extensive coverage of sensors over transportation networks, fueled by data processing technologies and computational power, has created an opportunity to build big data approaches; third, data-driven approaches do not rely on assumptions, hence more applicable for real-world problems. Data-driven methods have been applied in numerous predictive modeling such as traffic speed prediction [13], [39]–[42], travel time prediction [43], traffic flow prediction [44]–[47], vehicular queue length prediction [48], [49].

However, majority of the early studies explored traditional data-driven models such as, Support Vector Machine (SVM) [50], [51], $K$-Nearest Neighbor (KNN) [52]–[57], and Artificial Neural Network (ANN) [58]–[61] for traffic prediction. These studies are limited in scope and applicable for traffic prediction at a smaller spatial scale such as for a roadway segment or intersection. In recent years, many studies have applied deep neural network based approaches such as Long Short Term Memory Neural Network (LSTM), Convolutional Neural Network (CNN), Graph convolutional Neural Network (GCN) and integrated version of such models such as Convolutional LSTM (ConLSTM), Graph Convolutional LSTM (GCNLSTM) [10], [42], [62]–[65] for network scale traffic prediction. However, these studies mainly focused on learning traffic representation at a network scale rather than modeling network flow dynamics – how travel time variation impacts traffic flow propagation inside a network. Although a few studies [9], [10], [65] applied graph convolution based approach to dynamically learn spatiotemporal features in a transportation network, they are also limited to exploring different graph representation techniques such as latent network Laplacian matrix, constructing roadways inflow outflow probability matrix and weighted incident matrix to represent dynamic traffic patterns. In our study, we develop a method to learn the dynamics of a transportation network by capturing the correlation between travel time variations and traffic flow pattern of a transportation network.

Moreover, developing such large-scale model requires extensive data making them unsuitable for traffic management under emergencies such as incident or evacuation traffic management. However, to the best of our knowledge, none of the previous studies addressed the transferability challenge of such models for emergency management. In the study, to address this challenge we developed a method to transfer these models for real world traffic prediction during emergency event such as evacuation period. In addition, in case of evacuation traffic, it is critical to predict traffic well ahead of time (e.g., >1 hr.) to provide transportation agencies enough time to deploy traffic management strategies (i.e., signal control, emergency shoulder use etc.). However, to accurately predict traffic for a longer time period (>1 hr.) we need to account for demand variations over different time periods. Most of the existing studies disregard this concept as well. So, in our proposed method we incorporate both evacuation traffic demand related features with regular traffic features to sequentially predict evacuation traffic flow for a long-term horizon (1 to 6 hours).

## 3. Problem Formulation

To implement the method, we construct a network of traffic detectors where each detector indicates a node. In this network, travel time between two nodes dynamically changes over time. To capture the dynamics, we define the network as a dynamic graph $\mathcal{G}_t(v, \mathcal{E}, A_t)$ where $v$ denotes the set of nodes (i.e., detectors) and $\mathcal{E}$ denotes the set of links between nodes $(i, j)$. $A_t$ represents the connectivity between nodes as a weighted adjacency matrix, where weights are based on travel time between any two nodes $(i, j)$, defined as follows:

$$A_t(i, j) = \begin{cases} tt_{i,j}^t & if\ i \neq j \\ 0, & otherwise \end{cases} \tag{1}$$

where, $tt_{i,j}^t$ denotes the travel time between the nodes $i$ and $j$ at time $t$. The connectivity inside an adjacency matrix detects which neighboring nodes $(j)$ will be influenced by the traffic condition at a given node $(i)$. Moreover, in a time series problem the existing traffic condition at a given node $(i)$ will also influence its future traffic condition, which means each node is temporally self-influenced. This is represented by adding an identity $(I)$ matrix with the adjacency matrix which ensures that nodes are self-accessible,

$$\overline{A_t} = A_t + I \tag{2}$$

We aim to *learn* traffic flow patterns in a transportation network over multiple time steps (i.e., future time series) based on capturing the influence of congestion propagation (i.e., travel time variations) on spatiotemporal cross correlation among nodes' traffic condition. In this problem, traffic condition is represented as a function of traffic demand related features. Thus, we feed the model with the information on two aspects: (i) a dynamic graph indicating the variations in travel time and (ii) node level features related to traffic demand. Let, $X_t$ be the input features and $\mathcal{G}_t(v, \mathcal{E}, A_t)$ is a dynamic graph with weighted adjacency matrix $A_t$. The problem is defined as to learn a function $\mathcal{F}(.)$ that maps $l$ instances of input sequence $([X_{t-l}, X_{t-l+1} \dots, X_t])$ to predict $p$ instances of flow $(F_{t+1}, F_{t+2} \dots \dots \dots F_{t+p})$ for the entire network. Mathematically, the problem is defined as follows:

$$\mathcal{F}([X_{t-l}, X_{t-l+1} \dots, X_t]; [\mathcal{G}_{t-l}(v, \mathcal{E}, A_{t-l}) = [F_{t+1}, F_{t+2} \dots F_{t+p}] \tag{3}$$

where, $v\ and\ \mathcal{E}$ indicate the set of nodes and the set of links of the network, respectively; $l(= 0,1,2, \dots, l)$ and $p(= 1,2,3, \dots, p)$ indicate the input and output sequence, respectively; $X_t$ indicates traffic related features (i.e. volumes, time periods etc.); $A_t$ indicates the weighted adjacency matrix at time $t$; and the vector $F_{t+p}$ indicates the link flows for each link of the network at time $(t + p)$. We have added the description of the notations associated with the model development in Table 1.

## 4. Methodology

### 4.1 Learning Traffic Flow Dynamics of the Transportation Network

Traffic flow dynamics in a transportation network can be represented as a flow propagation process—traffic traversing from the origin node to the destination node via neighboring nodes. That is why, traffic condition of a given node influences the traffic condition of the neighboring nodes, in other words there exists a spatial correlation among these nodes. However, at any time step whether the traffic at given a node will reach any neighboring node or not depends on the travel time between these nodes, which changes over time. So, to model the traffic flow dynamics we need to represent travel time variations of the network and utilize this information while capturing spatial correlation among the nodes.

**Table 1** Description of the Notations Associated with the Model Development

| Notation | Description |
| --- | --- |
| $\mathcal{G}$ | Transportation network |
| $\boldsymbol{v}$ | Set of nodes in $\mathcal{G}$ with size of $\|v\| = N$ |
| $\mathcal{E}$ | Set of links in $\mathcal{G}$ with size of $\|\mathcal{E}\| = E$ |
| $\boldsymbol{A_t} \in \boldsymbol{R}^{N \times N}$ | Weighted adjacency matrix of $\mathcal{G}$, defined by Equation (1) |
| $\mathbf{I} \in \boldsymbol{R}^{N \times N}$ | Identity matrix |
| $\overline{\boldsymbol{A}}_t \in \boldsymbol{R}^{N \times N}$ | Neighborhood matrix defined by Equation (2) |
| $\overline{\boldsymbol{D}}_t \in \boldsymbol{R}^{N \times N}$ | Degree matrix of $\mathcal{G}$, a diagonal matrix where diagonal elements $(i, i)$ indicate the number of links coming out from a node |
| $\boldsymbol{tt_{ij}}$ | travel time between nodes $i$ and $j$ |
| $\boldsymbol{l}$ | Input time sequence length ( $0, 1, \ldots \ldots l$ ) |
| $\boldsymbol{c}$ | Number of features at each node |
| $\boldsymbol{X_t^{reg}} \in \boldsymbol{R}^{N \times c}$ | Contains all the traffic features (i.e. volumes, time periods etc.) associated with each node ($i$) of the network for regular condition |
| $\boldsymbol{X_t^{evc}} \in \boldsymbol{R}^{N \times c}$ | Contains all the traffic features (i.e. volumes, time periods etc.) associated with each node ($i$) of the network for evacuation condition |
| $\boldsymbol{X_t^{eD}} \in \boldsymbol{R}^{N \times c}$ | Contains the features related to evacuation travel demand (i.e population under mandatory, evacuation zones' location etc.) associated with each node ($i$) of the network |
| $\boldsymbol{g_t}$ | Graph Convolutional filter to learn the congestion propagation inside the network |
| $\boldsymbol{f}(.)$ | Activation function |
| $\mathbf{W_{gc}} \in \boldsymbol{R}^{N \times N}$ | Learnable parameters for the convolution filter |
| $\mathbf{H}$ | Indicates the outputs from different layers of the proposed neural network architecture |
| $\boldsymbol{p}$ | Prediction horizon ( $1, \ldots \ldots p$ ) |
| $\boldsymbol{F_{t+p}} \in \boldsymbol{R}^E$ | Flow vector contains flows for each link (segment) of the network for the prediction horizon |

All the bold letters denote a matrix

In this study, we develop a graph convolution based deep neural network architecture to capture spatiotemporal correlation among node-level traffic features for predicting traffic flows. The model has two layers (see Fig. 1): in the first layer, we apply a graph convolution operation to capture the spatial correlation among neighboring nodes. In this approach, we derive a graph convolutional filter from adjacency matrix which represents the travel time variations of the network, thereby detects which neighboring nodes are within the shortest path distance of the origin node at a given time step. To derive the convolutional filter, we adopt a graph theoretic approach where a graph adjacency matrix is decomposed into its eigenvalues to represent the structural properties of the graph such as the strength of a node (i.e., node level features), shortest path between two nodes etc. Such a representation, when fed into a deep learning model, suffers from exploding or vanishing gradient problem due to sparsity in eigen values' distribution. To overcome this exploding or vanishing gradient problem, Kipf and Welling [66] proposed a normalization technique to represent a graph and its intrinsic dynamics. We adopt a similar approach and define the graph as a symmetrically normalized adjacency matrix $(\overline{D}_t^{-\frac{1}{2}}\overline{A}_t\overline{D}_t^{-\frac{1}{2}})$. However, in previous applications the networks are static, hence normalized adjacency is fixed; while in our case the network is dynamic, hence the normalized adjacency matrix will change over time.

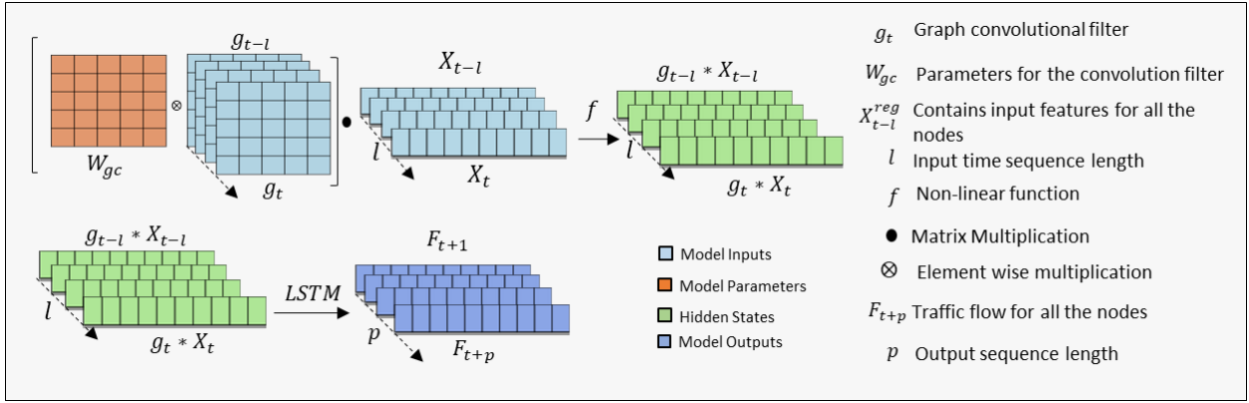

Fig. 1. A dynamic graph learning approach for network-wide traffic prediction

The main function of the graph convolution layer is to capture spatial cross correlation among nodes considering network wide travel time variations. We use the normalized graph adjacency matrix as a convolution filter and perform the convolution with node level traffic demand related features of the network. The convolution operation can be defined as follows:

$$gc_t = (W_{gc}\odot(\overline{D}_t^{-\frac{1}{2}}\overline{A}_t\overline{D}_t^{-\frac{1}{2}}))X_t \qquad (4)$$

where, $gc_t$ indicates the convoluted feature matrix and $W_{gc}$ indicates the parameters for the convolution filter. The convoluted feature matrix represents the state transition of the network, in other words how congestion is propagating inside the network and influencing neighboring nodes.

In the second layer, we apply a LSTM [67] model to map this convoluted feature matrix into traffic flows. The LSTM model captures the temporal dependency among traffic features while predicting traffic flows over multiple time steps (i.e., future time series). The proposed dynamic graph-based LSTM model (*DGCN-LSTM*) model can be defined as follows:

$$F_{t+p} = \text{LSTM}\big(f(gc_t)\big) = \text{LSTM}(f(W_{gc}\odot(\overline{D}_t^{-\frac{1}{2}}\overline{A}_t\overline{D}_t^{-\frac{1}{2}}))X_t) \qquad (5)$$

where, $f$ indicates the nonlinear activation function; we use rectified linear unit (relu) as an activation function.

Training such graph-based models over a transportation network requires a substantial amount of data. However, evacuations usually take place for 2 to 5 days before the landfall of a hurricane. When a model is trained with a small sample it may cause the model to overfit. To overcome this problem, we adopt a transfer learning technique [68]. We first train the *DGCN-LSTM* model over regular traffic data ($X_t = X_t^{reg}$) and later transfer this model to evacuation period. The following section describe the methods to implement the model for evacuation traffic prediction.

### 4.2 Network-wide Evacuation Traffic Prediction

Evacuation traffic depends on many factors such as zonal level population under mandatory evacuation, hours left before landfall, distance of a detector from the nearest evacuation zone, and different time periods of the day etc. [34]. Using these features as inputs, we can develop a simple time series-based model to predict evacuation traffic flow. However, such a model will perform poorly since it cannot capture the spatiotemporal dependency of traffic variables as it does not have any information on the underlying contexts of congestion propagation inside the network. To overcome this problem, we adopt a transfer learning approach to transfer the context of network dynamics over multiple time steps (temporal sequences). However, traffic demands during evacuation significantly differ from non-evacuation condition. For example, evacuation traffic demand is higher than non-evacuation period and does not follow any regular pattern. Thus, when applying the transfer learning approach, we need to transfer only the information relevant for an evacuation period such as information of network connection and the function how traffic flows from an upstream to a downstream location.

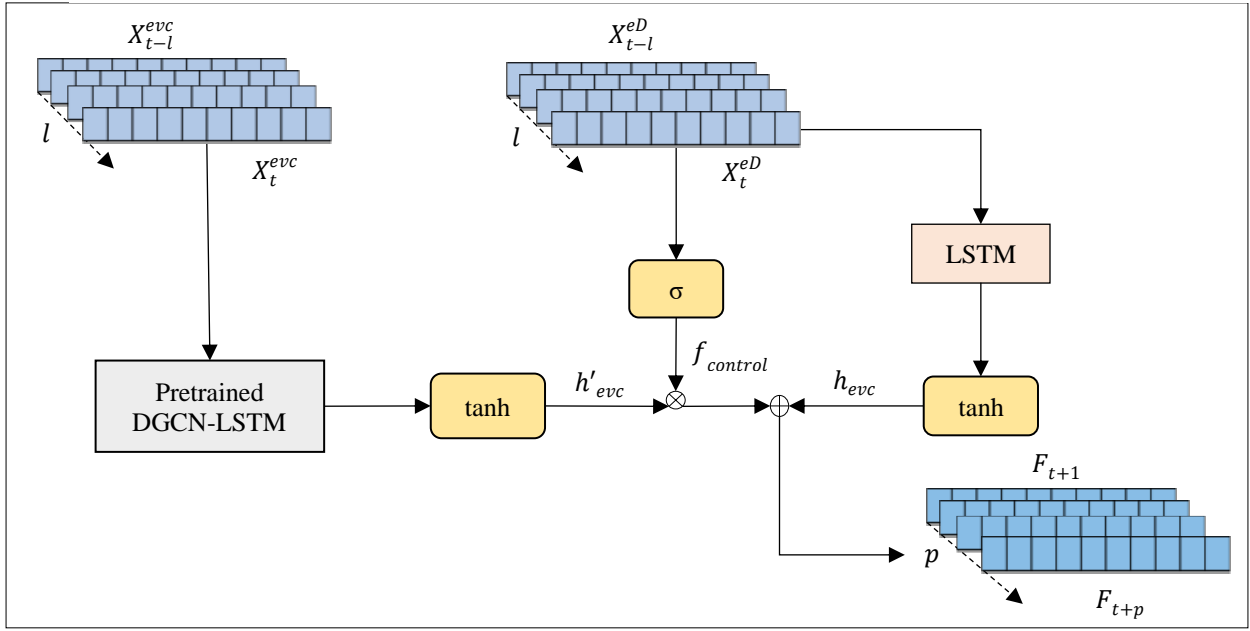

Fig. 2. A deep learning architecture for evacuation traffic prediction

We develop a deep learning architecture which controls the information flow from regular traffic condition to evacuation traffic condition. The proposed deep learning architecture has four components (see Fig. 2). The first component is the *pretrained DGCN-LSTM* model, we apply this model using the traffic features (i.e., volumes, time periods etc.) from evacuation period ($X_t = X_t^{evc}$) to predict traffic flow over multiple timesteps.

$$h'_{evc} = DGCNLSTM(X_t^{evc}) \qquad (6)$$

where, $h'_{evc}$ indicates the outputs from the *DGCN-LSTM* model. We add this information with other evacuation traffic demand information. The second component is an LSTM layer, we apply this model to capture the temporal correlation among evacuation traffic demand related features $(X_t^{eD})$.

$$h_{evc} = LSTM(X_t^{eD}) \tag{7}$$

where, $h_{evc}$ indicates the output from the LSTM layer. The third component is the control layer; in this layer, we define a neural network with sigmoid activation function to remove irrelevant information from the *DGCN-LSTM* model.

$$f_{control} = \sigma(W_C . X_t^{eD} + b_c) \tag{8}$$

where, $f_{control}$ indicates the output from the control layers which are distributed between 0 to 1.

The fourth and final component is the output layer which adds the network dynamics related information with evacuation demand to generate the final traffic prediction.

$$F_{t+p} = f_{control} \otimes h'_{evc} + tanh(h_{evc}) \tag{9}$$

In this layer, we perform an elementwise matrix multiplication between $f_{control}$ and $h'_{evc}$, thus some of the information will be erased prior to adding with evacuation demand. Since we assign weight $W_C$ in the control layer, when training the model for evacuation traffic prediction it automatically learns to control the information flow from non-evacuation condition to evacuation condition.

## 5. Data Collection and Preprocessing

### 5.1 Traffic Detector Data

To test the model, we consider a network consists of interstates highways. One of our objectives is to apply the model to predict evacuation traffic. Hence, to select the network we have observed previous evacuations patterns, which show that a large portion of residents living in Florida evacuates to Georgia or adjacent States. Thus, two major highways I-75, I-95 and other two highways I-4 and Florida Turnpike connecting them are expected to serve a substantial amount of evacuation traffic during Hurricane Irma. We choose the northbound direction of I-95, I-75, Florida Turnpike, and eastbound direction of I-4 (Fig. 3) to formulate the network.

We have collected data from Regional Integrated Transportation Information System (RITIS) [69] from September 4, 2017 to September 9, 2017 which covers the evacuation period of Hurricane Irma. We have also collected non-evacuation period traffic volume from May 1 to August 31, 2017. RITIS gathers data from Microwave Vehicle Detection System (MVDS) detectors deployed by the Florida DOT, giving real-time information on traffic speed, volume, and occupancy at a very high resolution (20 to 30s frequency).

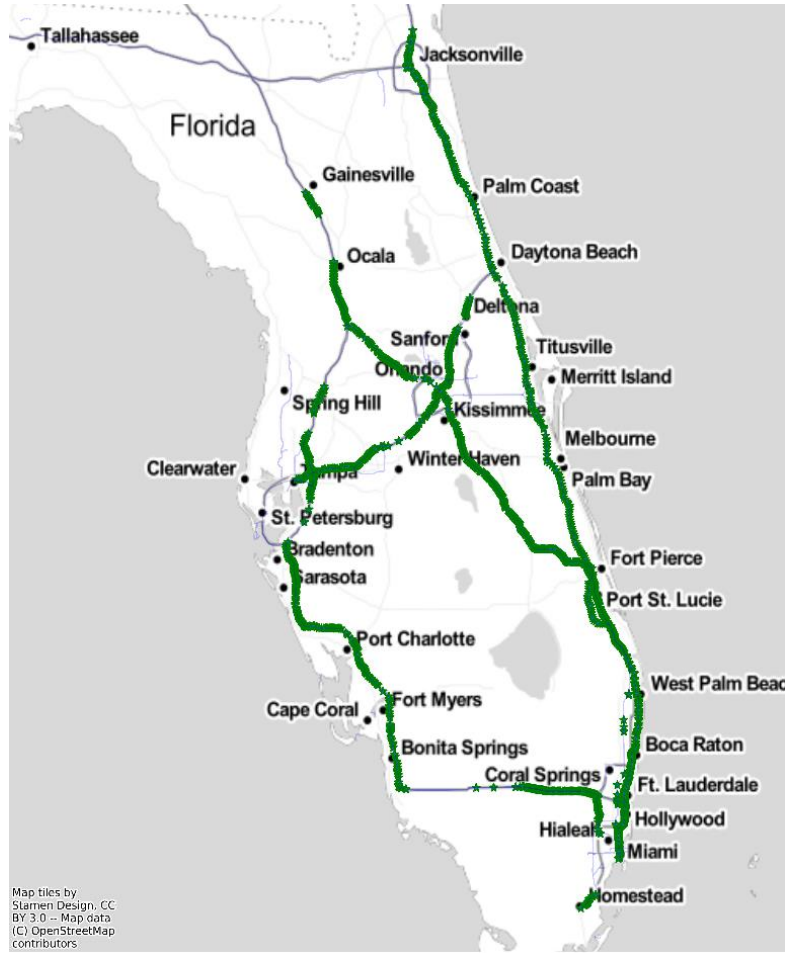

Fig. 3. Network of inter-state highways

The raw data collected from traffic detectors are subjected to errors. Several factors such as detector's malfunctioning, false encoding during storing the data into the server, overlapping of multiple entries, duplicate entries, bad weather conditions etc. can cause errors. Moreover, during congested stop and go traffic conditions, sometimes microwave radar detectors fail to detect vehicles, hence provide misleading information. Therefore, before proceeding to any data analysis, we need an extensive data cleaning and quality checking. Fig. 4 shows the framework for the data processing steps.

We followed several steps to process the data for analysis. Firstly, we remove the detectors having higher percentages of missing values (>20%); secondly, we detect the outliers based on the capacity of the highway (2500 vehicle per hour per lane); finally, replace the outliers and available missing values using multivariate iterative data imputation technique [70], [71]. The details about the data pre-processing are provided in our previous publication [34].

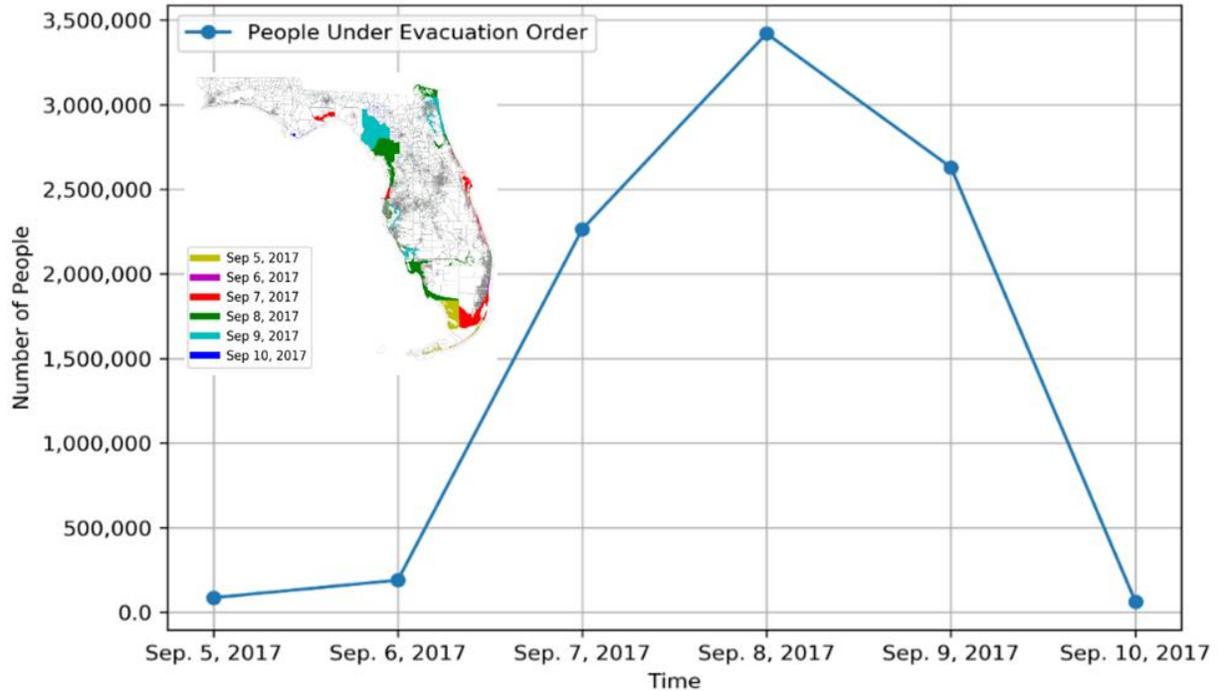

Fig. 4. Temporal variations of total population under mandatory evacuation

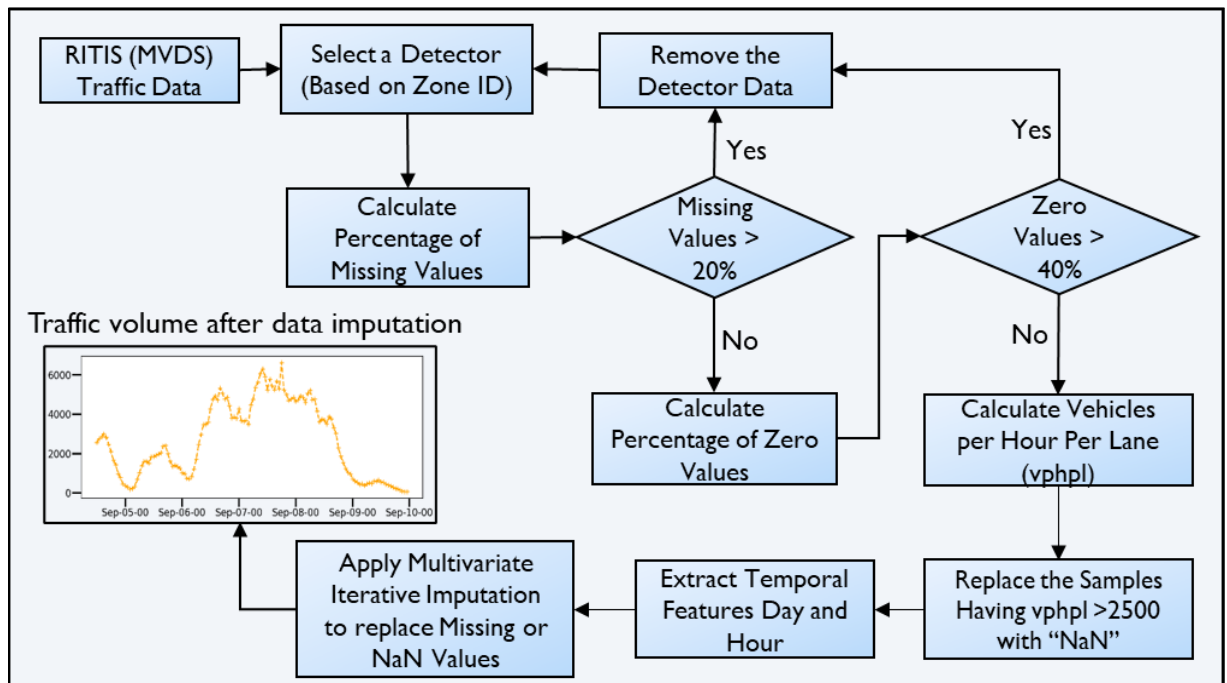

Fig. 5. Framework for data processing

## 5.2 Zonal Level Mandatory Evacuation

We collect the time and location of evacuation orders issued for different areas for Hurricane Irma from the Florida Division of Emergency Management. However, the declaration dates of evacuation order for all the zones are not available in a single source, thereby, in a few cases, we collect the declaration date by manually checking the emergency management agency's social media posts (e.g., Twitter, Facebook) of the respective county and contemporary news article available online. Fig. 5 shows the mandatory evacuation zones with declaration time. We observe that most of the evacuation zones are by the coast; smaller zones in the central part of Florida mainly represent mobile homes or low-lying areas vulnerable to inland flooding. Florida Keys and other low-lying zones such as Everglades were issued mandatory order in early September 5, 2017. Evacuation zones in the east coast, such as Miami-Dade, Daytona were issued evacuation order on September 7, 2017 (Hurricane Irma was supposed to hit the east coast of Florida until Sep. 7, 2017). After September 7, 2017, as the projected path shifted from east coast to west coast, evacuation zones of Naples, Cape Corals, Tampa, Levy, Jacksonville, were ordered mandatory evacuation on September 7, 2017, and onward (see Fig. 5). We have collected population data for the mandatory evacuation zones to understand how many people were under mandatory evacuation order. Since, population data is not available for the evacuation zones, we collect block group level population data from 2017 5-year American Community Survey and sum the population that falls within an evacuation zone to retrieve the population for the zone. The highest number of people were under a mandatory evacuation order on Sept. 8, 2017 (about 3,420,271 people), followed by Sept. 9, 2017 (about 2,629,161 people).

## 6. Experiments

### 6.1 Feature Extraction and Graph Representation

We followed several steps to extract the spatiotemporal features from the collected data. We prepared two types of data samples: (i) traffic data samples for regular period ($X_t^{reg}$) and evacuation period ($X_t^{evc}$); (ii) evacuation demand features ($X_t^{eD}$). All the input features are listed in Table 2.

### 6.1.1 Traffic data samples ($X_t$)

We aggregate the traffic data for 1-hour intervals estimating traffic flow and average traffic speed. To capture the periodic nature of traffic flow variations, we group the hours into 6 different time periods such as late night, early morning, morning, noon, evening, and night. We represent these features using one hot encoding, which means each of the six time periods is represented as an indicator variable (0,1). We also extract different features to present traffic flow variations over previous day ($t_{d-1}$) and previous time period ($t_{prd-1}$) corresponding to current day ($t_d$) and time periods ($t_{prd}$) at time ($t$). The extracted features include previous day and time periods' mean and standard deviation of traffic flow (see Table 2). Since we do not have any data to indicate the characteristics of different zones (e.g., built environment characteristics, zonal level population etc.); we use a variable named "Zone ID" to represent zonal characteristics specific to the location of each detector. This variable also represents the ordering of the output sequence (i.e., 1 to 806) for all the detectors. We formulate the traffic data sample as [number of samples ($n$), input time sequence ($l$), number of nodes ($N$), input features ($c$)]. Since we have collected the data from 806 detectors, the number of nodes, $N = 806$. We select 6-hour input data sequence to predict traffic for next 6 hour, so input time sequence length, $l = 6$ and prediction horizon length, $p = 6$. In total we have $c = 12$ input features: Zone ID, Late Night (12am-4am), Early Morning (4am-8am), Morning (8am -12 pm), Mid-day (12 pm-4pm), Evening (4pm-8pm), Night (8pm -12am), mean traffic flow ($q_t$), previous day mean traffic flow ($\overline{q}_{t_{d-1}}$), previous day standard deviation of traffic flow ($sq_{t_{d-1}}$), previous time period mean traffic flow ($\overline{q}_{t_{prd-1}}$), previous time period standard deviation in traffic flow ($sq_{t_{prd-1}}$). For non-evacuation period, we have the data for 2148 hours and for evacuation period we have data for 120 hours. So, for evacuation and non-evacuation periods the input data has the shape as [120, 6, 806, 12] and [2148, 6, 806, 12], respectively and the target data has the shape as [120, 6, 806] and [2148, 6, 806], respectively.

**Table 2** Description of Input Features

| Variables | Description |
|---|---|
| $Z$ | Zone ID: represent zonal characteristics specific to the location of each detector (i.e. 1,2,….N) |
| $t_{prd}$ | Time periods: Late Night (12am-4am), Early Morning (4am-8am), Morning (8am -12 pm), Mid-day (12 pm-4pm), Evening (4pm-8pm), Night (8pm -12am) |
| $q_t$ | Traffic flow at time t |
| $\overline{q}_{t_{d-1}}$ | Previous day mean traffic flow |
| $sq_{t_{d-1}}$ | Previous day standard deviation of traffic flow |
| $\overline{q}_{t_{prd-1}}$ | Previous time period mean traffic flow |
| $sq_{t_{prd-1}}$ | Previous time period standard deviation of traffic flow |
| $S$ | Mean speed over an hour |
| $T_l$ | Hours left before landfall |
| $P_{evc}$ | Cumulative population under mandatory evacuation |
| $d_{evc}$ | Distance to the nearest evacuation zone from each detector |

*6.1.2   Evacuation demand data samples* $(X_t^{eD})$

We extract features related to evacuation demand such as population under mandatory evacuation zone, distance of a detector from the nearest evacuation zone, and hours left before hurricane landfall. In case of the variable "cumulative population under mandatory evacuation" we consider a time lag of 18 hr. between the declaration of evacuation order and the time when people start to evacuate. To account this time lag, we shift the total population under mandatory evacuation zone by 18 hr. We perform an empirical analysis by running linear regression model multiple times with different time lag and find that the coefficient associated with the variable total population under mandatory order is positive for 18-hour time lag and significantly influence the increase in evacuation traffic flow. The details on fixing the time lag is provided in [34]. Moreover, evacuation demand also depends on time period of the day. From our previous study [34], we find that people are more likely to evacuate during day time compare to night time. So, we also consider time periods to capture evacuation traffic demand. Finally, the evacuation demand data samples have 9 features: Late Night (12am-4am), Early Morning (4am-8am), Morning (8am -12 pm), Mid-day (12 pm-4pm), Evening (4pm-8pm), Night (8pm -12am), population under mandatory evacuation zone ($P_{evc}$), distance of the nearest evacuation zone from each detector ($d_{evc}$), and hours left before hurricane landfall $(T_l)$. We formulate the evacuation demand data as [number of samples $(n)$, input time sequence $(l)$, number of nodes $(N)$, input features $(c)$] i.e. [120, 6, 806, 9].

### 6.1.3 Graph Representation

we follow several steps to construct the graph. First, we map the detectors' locations into Open Street Map. Second, considering the detectors as nodes, we connect these detectors to complete the network. Finally, after constructing the network, we represent the network using adjacency matrix (see equation 1 and 2). We also calculate the travel distance between pairs of nodes from the open street map and estimate the travel time. We define the travel time between a pair of nodes as follows,

$$t_{ij} = \frac{d_{ij}}{\frac{S_i + S_j}{2}} = \frac{2d_{ij}}{S_i + S_j} \tag{10}$$

where, $t_{ij}, d_{ij}$ indicate the travel time and distance between two consecutive detectors; $S_i$ and $S_j$ indicate average speed for two consecutive detectors. We use the travel time as weight for the adjacency matrix. We also perform gaussian transformation on the weighted graph adjacency matrix,

$$A_t(i,j) = \begin{cases} \exp\left(-\frac{tt_{i,j}^2}{tt_{std}^2}\right), & \text{if } i \neq j \text{ and } \exp\left(-\frac{tt_{i,j}^2}{tt_{std}^2}\right) \geq r \\ 0, & otherwise \end{cases} \tag{11}$$

where, $tt_{std}$ indicate the standard deviation of travel time and $r$ is the threshold to control the distribution and sparsity of weighted graph adjacency matrix, we fix the threshold value as 0.1 based on previous studies [72] and experiments results.

## 6.2 Baseline Models

We implement three baseline models to compare the performance of the proposed DGCN-LSTM model.

### 6.2.1 LSTM

In the *LSTM* model we use two stacked *LSTM* layer to predict traffic for next 6 hour. Each of the layer we assign 4836 (number of nodes * output sequence length) hidden neurons. The output layer is a fully connected layer with tanh activation function.

### 6.2.2 Convolutional LSTM

In the Convolutional LSTM (*ConvLSTM*) model, we stack a convolution layer with an *LSTM* layer. The convolutional layer uses a convolution filter to extract the spatial correlation among traffic features between consecutive detectors. We experiment with different size of the kernel $(k)$ and find that the model performs best for a kernel size of 3. The output from the convolutional layer is fed into the *LSTM* layer to capture temporal correlation among traffic features while predicting traffic flow over a long sequence.

### 6.2.3 Graph Convolutional LSTM

In the graph convolutional LSTM *(GCN-LSTM)* model, we apply a similar approach as [66], [72]. In this case, the weights of the graph adjacency matrix are constant and assigned based on the distance between two consecutive nodes.

## 6.3 Model Training

We train the model using mean squared error as the loss function. At each iteration, the model estimates the mean squared error for the predicted flows $(\hat{F}_{t+p}^\varepsilon)$ and the actual flows $(F_{t+p}^\varepsilon)$ of the network. Afterward, the gradient of the loss function is backpropagated to adjust the weights to reduce loss function value. The loss function can be defined as,

$$L = Loss\left(F_{t+p}^{\mathcal{E}}, \hat{F}_{t+p}^{\mathcal{E}}\right) \tag{12}$$

$$MSE = \frac{1}{p}\sum_{p=1}^{p}\frac{\sum_{\mathcal{E}=1}^{E}(F_{t+p}^{\mathcal{E}} - \hat{F}_{t+p}^{\mathcal{E}})^2}{E} \tag{13}$$

where, $Loss(.)$ is the function to estimate the error between the actual ($F_{t+p}^{\mathcal{E}}$) and estimated values ($\hat{F}_m^{\mathcal{E}}$) and $\mathcal{E}$ denotes the set of links for the network. We implement our model using Pytorch environment [73] and train the model with dual NVIDIA Tesla V100 16GB PCIe GPU.

### 6.3.1 DGCN-LSTM for Non-evacuation period

From the regular traffic data samples, we use 90% for training, 5% for validation, and rest 5% of the data for testing the model. Based on the validation accuracy, we tune the hyperparameters such as learning rate, maximum number of iterations, and the type of the optimizer. We also track the training and validation loss values to check whether the model is overfitting or not. From the loss values, we find that it takes about 60 epochs with a learning rate of 0.001 for the model to converge (i.e., similar train and validation loss value). After that there are merely any variation in loss values (Fig. 6 (a)). Moreover, after 70 epochs the value of the loss function for the validation data gradually starts increasing, indicating that the model starts to overfit. We use Adaptive Moment Estimation (ADAM) to train the model. Compared to other optimizers such as Adaptive Gradient (AdaGrad), Root Mean Square Propagation (RMSProp) etc., ADAM optimizer gives more stable solutions.

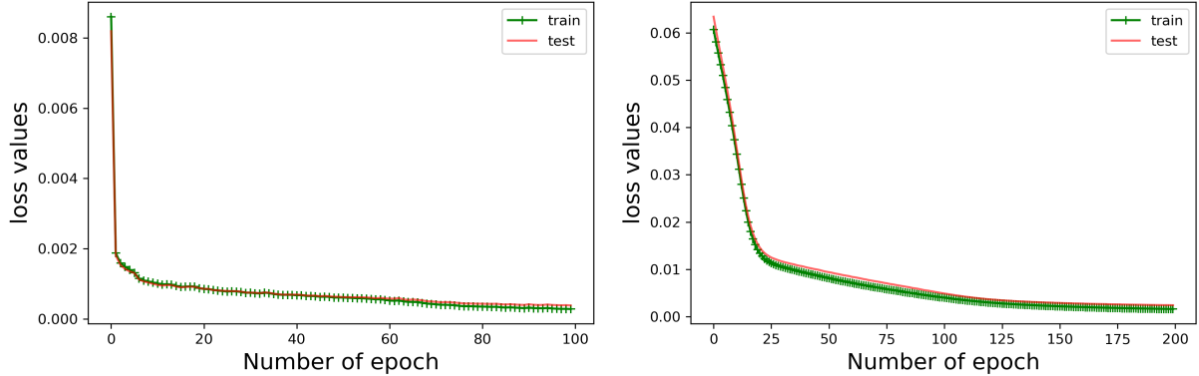

Fig. 6. Variations of training and validation loss (a) DGCN-LSTM (b) Transfer Learned DGCN-LSTM

### 6.3.2 Transfer Learned DGCN-LSTM for Evacuation period:

From the evacuation traffic data samples, we use 80% for training, 10% for validation, and rest 10% of the data for testing the model. Similar to previous model, we experimented with different optimizer however ADAM optimizer gives the best result. We also track the changes in training and validation loss values to ensure the model is not overfitting. It takes 150 epochs for the model to converge, after that it starts to overfit (Fig. 6(b)).

## 6.4 Experiment results

Once the final model is fixed, we test it on the test data set. We calculate Root Mean Squared Error (RMSE) and Mean Absolute Error (MAE) as performance measures to check the accuracy of the implemented model. Performance metrics are defined as:

$$RMSE = \sqrt{\frac{1}{p}\sum_{p=1}^{p}\frac{\sum_{\mathcal{E}=1}^{E}(F_{t+p}^{\mathcal{E}} - \hat{F}_{t+p}^{\mathcal{E}})^2}{E}} \tag{14}$$

$$\text{MAE} = \frac{1}{p}\sum_{p=1}^{p}\frac{\sum_{\mathcal{E}=1}^{E}|F_{t+p}^{\mathcal{E}} - \hat{F}_{t+p}^{\mathcal{E}}|}{E} \tag{15}$$

In Table 3, we report the performance of the model on test dataset. To account the sensitivity of the model over different data samples, we randomly split the data to generate 10 different train, test, and validation data sets. Finally, we train 10 different models and report the mean and standard deviation of the estimated performance measures on the test data sets. Based on performance measures, we find that the proposed *DGCN-LSTM* model performed best compared to other baseline models. The RMSE and MAE values of the model is 226.85 and 133.82, respectively. However, RMSE and MAE provide aggregate information (average over all the outputs) on the performance of the models. Hence, we also estimate the $R^2$ score. As shown in Table 3, the $R^2$ score for the proposed model is 0.98 which means the model can learn the regular traffic flow patterns very well.

**Table 3** Comparisons among different models to predict traffic over 6-hour sequence for non-evacuation period

| Model | Mean RMSE | Std RMSE | Mean MAE | Std MAE | Mean $R^2$ Score | Std $R^2$ Score |
|-------|-----------|----------|----------|---------|------------------|-----------------|
| LSTM | 282.38 | 14.24 | 160.81 | 3.92 | 0.98 | 0.0028 |
| GCN-LSTM | 275.08 | 14.08 | 152.08 | 5.91 | 0.98 | 0.0024 |
| ConvLSTM | 246.57 | 31.19 | 146.11 | 14.94 | 0.98 | 0.0046 |
| DGCN-LSTM | 226.84 | 21.54 | 133.82 | 9.58 | 0.98 | 0.0032 |

*std = standard deviation*

However, when we apply the model for evacuation traffic prediction it performs poorly; the RMSE and MAE values significantly increased to 1440.994 and 1009.94, respectively. Hence, we use the transfer learning approach with additional demand features to capture the changes in traffic demand during hurricane period, which improves the overall prediction accuracy. For the transfer learned model the RMSE and MAE values are 399.69 and 268.03, respectively (Table 4).

**Table 4** Performance of the proposed model for evacuation traffic prediction

| Model | Min flow | Max flow | RMSE | MAE | $R^2$ Score |
|-------|----------|----------|------|-----|-------------|
| DGCN-LSTM | 50.0 | 11,2470 | 1440.99 | 1009.94 | 0.21 |
| Transfer learned DGCN-LSTM | 50.0 | 11,2470 | 399.69 | 268.03 | 0.943 |

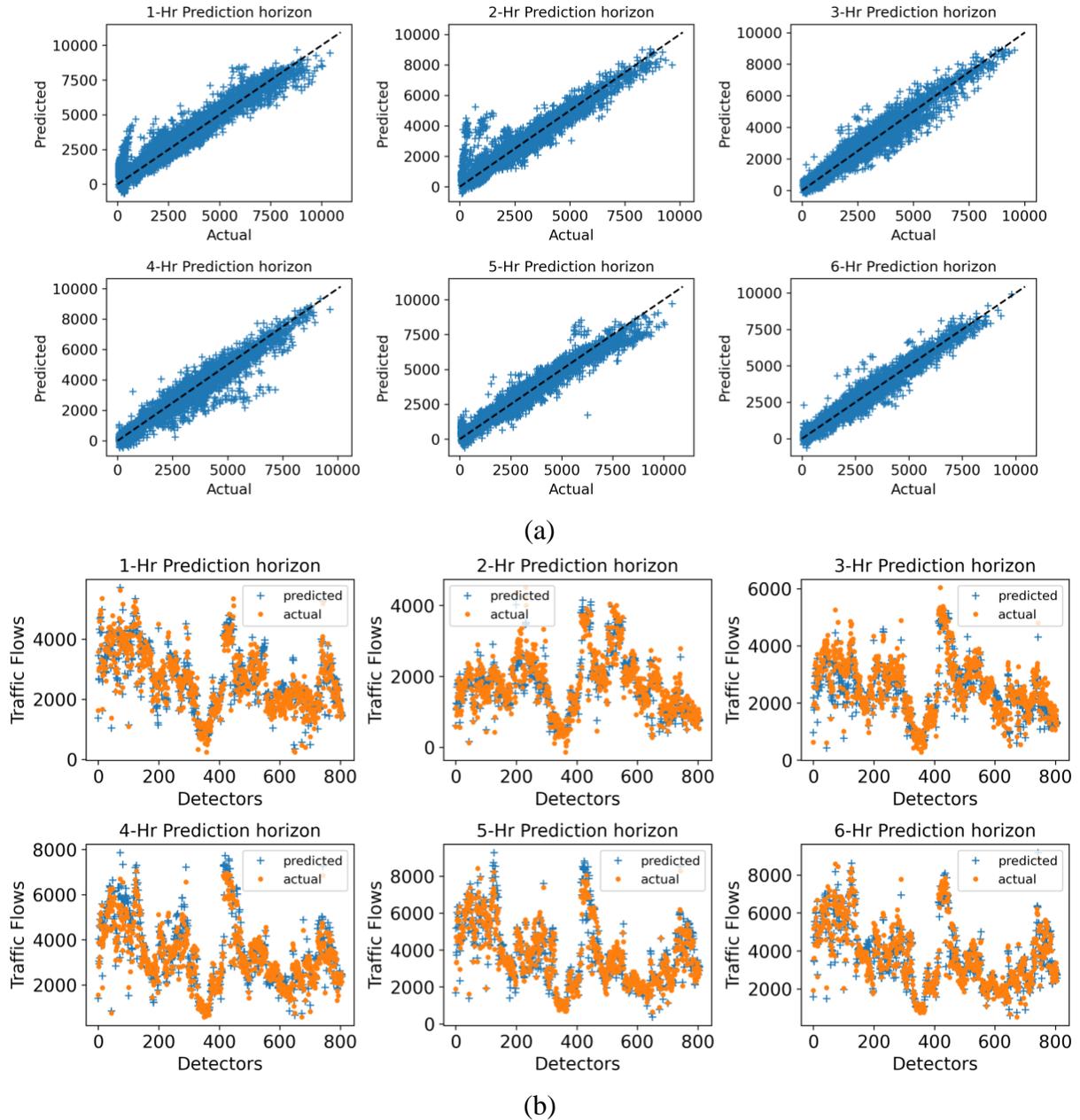

(a)

(b)

Fig. 7. Comparison between actual and predicted traffic flow (a) correlation (b) variations over different detectors

Fig. 7 (a) shows the relation between actual and predicted link flows, indicating that the actual and predicted traffic flow almost match with each other. We also find that in some cases the model overestimates the flows. This is because of the drop in overall traffic at the upstream locations just before the landfall day due to changes in hurricane path and mandatory evacuation zones. We also observe the detector wise variations of actual and predicted traffic flows (Fig.7 (b)), which indicates that the model can capture spatiotemporal patterns of traffic very well.

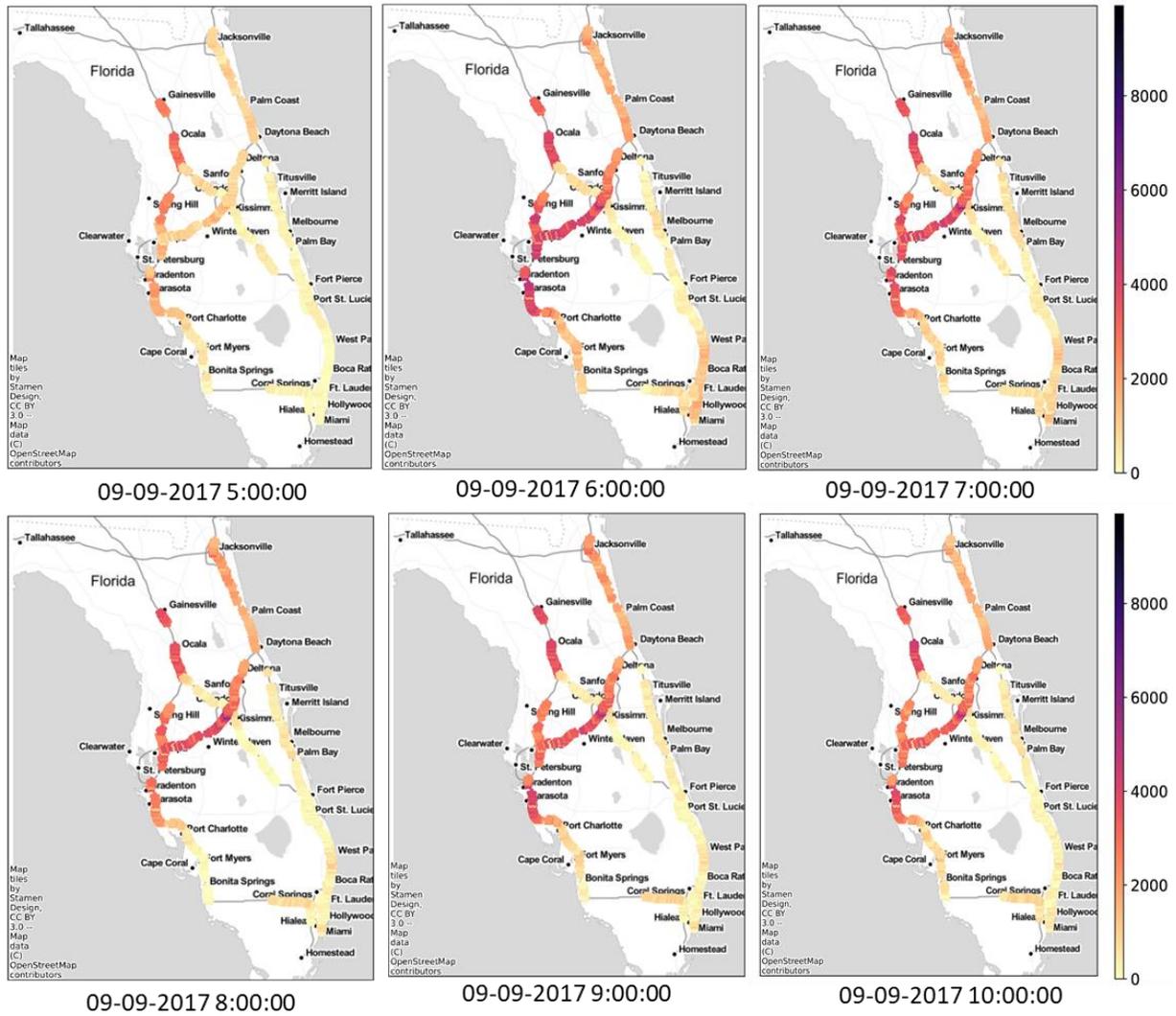

Fig. 8. Congestion map indicating traffic flow (predicted) variations over different time during hurricane evacuation

*6.5 Congestion Mapping to Understand Network Disruption*

We apply the implemented model to predict traffic for different days prior to hurricane landfall. As shown in Fig. 8, we map the predicted traffic flows to generate spatiotemporal traffic variations for different zones over a 6-hour period. The figure demonstrates congestion propagation at different zones of the network, such information will be critical for the emergency traffic management agencies to implement strategies focusing on reducing delay during hurricane evacuation. The figure provides further evidence that the implement model is able to capture the spatiotemporal traffic variations of a network during emergency evacuation even in case of unexpected event such as changes in hurricane causing zonal level evacuation traffic demand. For example, we observe a significant traffic congestion near west coast of Florida region: downstream of I75 and I4 on September 8 and 9, 2017. This is because of shift in hurricane path which forced people living in Naples, Cape Corals, Tampa, Levy, Jacksonville to evacuate at the eleventh hour causing significant congestion.

## 7. Conclusions

Evacuation traffic prediction is one of the most critical elements for deploying pro-active traffic management strategies. However, with higher traffic volume and no peaking behavior, evacuation traffic patterns differ from non-evacuation traffic conditions. Thus, it is more challenging to learn such irregularities using traditional modeling approaches. Moreover, modeling spatiotemporal traffic variations requires large volume of data at a higher resolution, which is quite impossible to obtain due to the short span of an evacuation period. To address these challenges, we develop a new method considering spatiotemporal network dynamics to accurately predict evacuation traffic over multiple time steps.

First, we develop a deep learning architecture namely *DGCN-LSTM* to learn the spatiotemporal network scale traffic patterns and train the model with non-evacuation period traffic data. Based on the experiment results, we find that the implemented *DGCN-LSTM* outperforms the existing deep learning models such as *LSTM*, *ConvLSTM* and *GCNLSTM* model with an RMSE of 226.84. However, as we apply the model for evacuation period the RMSE value increased to 1440.99. We overcome this issue by adopting a transfer learning approach with additional evacuation traffic demand related features such as distance from the evacuation zone, time to landfall, and other zonal level features to control the information flow from the pretrained *DGCN-LSTM* model. The final transfer learned *DGCNLSTM* model performs well to predict evacuation traffic flow (RMSE 399.69).

This study has several limitations. We have not tested the performance of the model over data from other hurricanes. Testing the model for multiple hurricanes will further establish the generalizability of the model for evacuation traffic prediction. Moreover, we do not have data related to zonal level evacuation traffic demand, we use aggregate measures such as total population under mandatory evacuation to understand overall demand. High resolution demand data from emerging technologies such as mobile phone sensors, connected vehicles data can be used to overcome this issue.

### ACKNOWLEDGMENT

This study was supported by the U.S. National Science Foundation through the grants CMMI #1917019 and CMMI #1832578. However, the authors are solely responsible for the facts and accuracy of the information presented in the paper.

### REFERENCES

[1]     F. Report, "Select Committee on Hurricane Response and Preparedness," 2018.

[2]     P. Murray-Tuite and B. Wolshon, "Evacuation transportation modeling: An overview of research, development, and practice," *Transp. Res. Part C Emerg. Technol.*, vol. 27, no. 1882, pp. 25–45, 2013.

[3]     P. Murray-Tuite, B. Wolshon, and D. Matherly, "Evacuation and Emergency Transportation: Techniques and Strategis for Systems Resilience," *TR NEWS 311*, D.C., Washington, 2017.

[4]     B. Barrett, B. Ran, and R. Pillai, "Developing a dynamic traffic management modeling framework for hurricane evacuation," *Transp. Res. Rec.*, no. 1733, pp. 115–121, 2000.

[5]     Q. Li, X. Yang, and H. Wei, "Integrating traffic simulation models with evacuation planning system in a GIS environment," *IEEE Conf. Intell. Transp. Syst. Proceedings, ITSC*, pp. 590–595, 2006.

[6]     Y. Chen, S. Y. Shafi, and Y. fan Chen, "Simulation pipeline for traffic evacuation in urban areas and emergency traffic management policy improvements through case studies," *Transp. Res. Interdiscip. Perspect.*, vol. 7, 2020.

[7]     B. Staes, N. Menon, and R. L. Bertini, "Analyzing transportation network performance during emergency evacuations: Evidence from Hurricane Irma," *Transp. Res. Part D Transp. Environ.*, vol. 95, no. April, p. 102841, 2021.

[8]     M. Ghorbanzadeh, S. Burns, L. V. N. Rugminiamma, E. Erman Ozguven, and W. Huang, "Spatiotemporal Analysis of Highway Traffic Patterns in Hurricane Irma Evacuation," *Transp. Res. Rec. J. Transp. Res. Board*, p. 036119812110018, 2021.

[9]     H. Peng *et al.*, "Spatial temporal incidence dynamic graph neural networks for traffic flow forecasting," *Inf. Sci. (Ny).*, vol. 521, pp. 277–290, 2020.

[10]    H. Peng *et al.*, "Dynamic graph convolutional network for long-term traffic flow prediction with reinforcement learning," *Inf. Sci. (Ny).*, vol. 578, pp. 401–416, 2021.

[11]    K. Guo, Y. Hu, Z. Qian, Y. Sun, J. Gao, and B. Yin, "Dynamic Graph Convolution Network for Traffic Forecasting Based on Latent Network of Laplace Matrix Estimation," *IEEE Trans. Intell. Transp. Syst.*, pp. 1–10, 2020.

[12]    Z. Cui, R. Ke, Z. Pu, and Y. Wang, "Stacked bidirectional and unidirectional LSTM recurrent neural network for forecasting network-wide traffic state with missing values," *Transp. Res. Part C Emerg. Technol.*, vol. 118, no. March, p. 102674, 2020.

[13]    Z. Cui, K. Henrickson, R. Ke, and Y. Wang, "Traffic Graph Convolutional Recurrent Neural Network: A Deep Learning Framework for Network-Scale Traffic Learning and Forecasting," *IEEE Trans. Intell. Transp. Syst.*, pp. 1–12, 2019.

[14]    P. Murray-Tuite, M. K. Lindell, B. Wolshon, and E. J. Baker, *Large-Scale Evacuation: The Analysis, Modeling, and Management of Emergency Relocation from Hazardous Areas*. CRC Press, 2018.

[15]    S. Wong, S. Shaheen, and J. Walker, "Understanding evacuee behavior: A case study of hurricane Irma," 2018.

[16]    G. P. Moynihan and D. J. Fonseca, "Design of an Evacuation Demand Forecasting Module for Hurricane Planning Applications," *J. Transp. Technol.*, vol. 06, no. 05, pp. 257–276, 2016.

[17]    A. M. Sadri, S. V. Ukkusuri, and P. Murray-Tuite, "A random parameter ordered probit model to understand the mobilization time during hurricane evacuation," *Transp. Res. Part C Emerg. Technol.*, vol. 32, no. November 2015, pp. 21–30, 2013.

[18]    S. Hasan, S. Ukkusuri, H. Gladwin, and P. Murray-Tuite, "Behavioral Model to Understand Household-Level Hurricane Evacuation Decision Making," *J. Transp. Eng.*, vol. 137, no. 5, pp. 341–348, 2011.

[19]    S. Hasan, R. Mesa-Arango, and S. Ukkusuri, "A random-parameter hazard-based model to understand household evacuation timing behavior," *Transp. Res. Part C Emerg. Technol.*, vol. 27, no. January, pp. 108–116, 2013.

[20]    S.-K. Huang, M. K. Lindell, and C. S. Prater, "Who leaves and who stays? A review and statistical meta-analysis of hurricane evacuation studies," *Environ. Behav.*, vol. 48, no. 8,

pp. 991–1029, 2016.

[21] J. Fry and J. M. Binner, "Elementary modelling and behavioural analysis for emergency evacuations using social media," *Eur. J. Oper. Res.*, vol. 249, no. 3, pp. 1014–1023, 2015.

[22] R. Gudishala and C. Wilmot, "Predictive Quality of a Time-Dependent Sequential Logit Evacuation Demand Model," *Transp. Res. Rec. J. Transp. Res. Board*, vol. 2376, pp. 38–44, 2013.

[23] T. Rambha, L. Nozick, and R. Davidson, "Modeling Departure Time Decisions During Hurricanes Using a Dynamic Discrete Choice Framework," 2019.

[24] A. J. Pel, M. C. J. Bliemer, and S. P. Hoogendoorn, "A review on travel behaviour modelling in dynamic traffic simulation models for evacuations," *Transportation (Amst).*, vol. 39, no. 1, pp. 97–123, 2012.

[25] R. Mesa-arango, S. Hasan, S. V Ukkusuri, A. M. Asce, and P. Murray-tuite, "Household-Level Model for Hurricane Evacuation Destination Type Choice Using Hurricane Ivan Data," *Nat. Hazards Rev.*, vol. 14, no. February, pp. 11–20, 2013.

[26] C. G. Wilmot, N. Modali, and B. Chen, "Modeling Hurricane Evacuation Traffic: Testing the Gravity and Intervening Opportunity Models as Models of Destination Choice in Hurricane Evacuation," 2006. [Online]. Available: https://rosap.ntl.bts.gov/view/dot/22136.

[27] R. A. Davidson *et al.*, "An Integrated Scenario Ensemble-Based Framework for Hurricane Evacuation Modeling: Part 1—Decision Support System," *Risk Anal.*, vol. 40, no. 1, pp. 97–116, 2020.

[28] K. Yang *et al.*, "Hurricane evacuations in the face of uncertainty: Use of integrated models to support robust, adaptive, and repeated decision-making," *Int. J. Disaster Risk Reduct.*, vol. 36, p. 101093, 2019.

[29] B. Blanton *et al.*, "An Integrated Scenario Ensemble-Based Framework for Hurricane Evacuation Modeling: Part 2—Hazard Modeling," *Risk Anal.*, vol. 40, no. 1, pp. 117–133, 2020.

[30] R. Gudishala and C. Wilmot, "Comparison of Time-Dependent Sequential Logit and Nested Logit for Modeling Hurricane Evacuation Demand," *Transp. Res. Rec.*, vol. 2312, no. 1, pp. 134–140, 2012.

[31] T. Litman, "Lessons from Katrina and Rita: What major disasters can teach transportation planners," *J. Transp. Eng.*, vol. 132, no. 1, pp. 11–18, 2006.

[32] V. Dixit and B. Wolshon, "Evacuation traffic dynamics," *Transp. Res. Part C Emerg. Technol.*, vol. 49, pp. 114–125, 2014.

[33] R. Rahman, S. Hasan, and M. H. Zaki, "Towards reducing the number of crashes during hurricane evacuation: Assessing the potential safety impact of adaptive cruise control systems," *Transp. Res. Part C Emerg. Technol.*, vol. 128, no. April, p. 103188, 2021.

[34] R. Rahman, K. C. Roy, and S. Hasan, "Understanding Network Wide Hurricane Evacuation Traffic Pattern from Large-scale Traffic Detector Data," in *IEEE Conference on Intelligent Transportation Systems, Proceedings, ITSC*, 2021, vol. 2021-Septe, pp. 1827–1832.

[35] R. Rahman and S. Hasan, "Short-Term Traffic Speed Prediction for Freeways during Hurricane Evacuation: A Deep Learning Approach," *IEEE Conf. Intell. Transp. Syst. Proceedings, ITSC*, vol. 2018-Novem, no. November 2019, pp. 1291–1296, 2018.

[36] K. C. Roy, S. Hasan, A. Culotta, and N. Eluru, "Predicting traffic demand during hurricane evacuation using Real-time data from transportation systems and social media,"

*Transp. Res. Part C Emerg. Technol.*, vol. 131, no. July, p. 103339, 2021.

[37]    R. Yu, Y. Li, C. Shahabi, U. Demiryurek, and Y. Liu, "Deep learning: A generic approach for extreme condition traffic forecasting," *Proc. 17th SIAM Int. Conf. Data Mining, SDM 2017*, pp. 777–785, 2017.

[38]    S. Oh, Y. J. Byon, K. Jang, and H. Yeo, "Short-term travel-time prediction on highway: A review on model-based approach," *KSCE J. Civ. Eng.*, no. April, pp. 1–13, 2017.

[39]    X. Ma, Z. Tao, Y. Wang, H. Yu, and Y. Wang, "Long short-term memory neural network for traffic speed prediction using remote microwave sensor data," *Transp. Res. Part C Emerg. Technol.*, vol. 54, pp. 187–197, 2015.

[40]    Z. Cui and Y. Wang, "Deep Stacked Bidirectional and Unidirectional LSTM Recurrent Neural Network for Network-wide Traffic Speed Prediction," pp. 22–25, 2017.

[41]    T. Epelbaum, F. Gamboa, J.-M. Loubes, and J. Martin, "Deep Learning applied to Road Traffic Speed forecasting," 2017.

[42]    L. Zhao *et al.*, "T-GCN: A Temporal Graph Convolutional Network for Traffic Prediction," *IEEE Trans. Intell. Transp. Syst.*, vol. 21, no. 9, pp. 3848–3858, 2020.

[43]    Yanjie Duan, Yisheng Lv, and Fei-Yue Wang, "Travel time prediction with LSTM neural network," *2016 IEEE 19th Int. Conf. Intell. Transp. Syst.*, pp. 1053–1058, 2016.

[44]    B. Yang, S. Sun, J. Li, X. Lin, and Y. Tian, "Traffic flow prediction using LSTM with feature enhancement," *Neurocomputing*, vol. 332, pp. 320–327, 2019.

[45]    X. Luo, D. Li, Y. Yang, and S. Zhang, "Spatiotemporal Traffic Flow Prediction with KNN and LSTM," vol. 2019, 2019.

[46]    N. G. Polson and V. O. Sokolov, "Deep learning for short-term traffic flow prediction," *Transp. Res. Part C Emerg. Technol.*, vol. 79, pp. 1–17, 2017.

[47]    K. Guo *et al.*, "Optimized Graph Convolution Recurrent Neural Network for Traffic Prediction," *IEEE Trans. Intell. Transp. Syst.*, vol. 22, no. 2, pp. 1138–1149, 2021.

[48]    R. Rahman and S. Hasan, "Real-time signal queue length prediction using long short-term memory neural network," *Neural Comput. Appl.*, vol. 33, no. 8, pp. 3311–3324, 2021.

[49]    S. Lee, K. Xie, D. Ngoduy, and M. Keyvan-Ekbatani, "An advanced deep learning approach to real-time estimation of lane-based queue lengths at a signalized junction," *Transp. Res. Part C Emerg. Technol.*, vol. 109, pp. 117–136, 2019.

[50]    C. Wu, C. Wei, D. Su, M. Chang, and J. Ho, "Travel time prediction with support vector regression," *Proc. 2003 IEEE Int. Conf. Intell. Transp. Syst.*, vol. 2, pp. 1438–1442, 2004.

[51]    J. Ahn, "Highway traffic flow prediction using support vector regression and Bayesian classifier," *2016 Int. Conf. Big Data Smart Comput.*, pp. 239–244, 2016.

[52]    F. G. Habtemichael and M. Cetin, "Short-term traffic flow rate forecasting based on identifying similar traffic patterns," *Transp. Res. Part C Emerg. Technol.*, vol. 66, pp. 61–78, 2016.

[53]    B. Yu, X. Song, F. Guan, Z. Yang, and B. Yao, "k-Nearest Neighbor Model for Multiple-Time-Step Prediction of Short-Term Traffic Condition," *J. Transp. Eng.*, vol. 142, no. 6, p. 04016018, 2016.

[54]    W. Qiao, A. Haghani, and M. Hamedi, "A Nonparametric Model for Short-Term Travel Time Prediction Using Bluetooth Data," *J. Intell. Transp. Syst.*, vol. 17, no. 2, pp. 165–175, 2013.

[55]    M. Meng, C. Shao, Y. Wong, B. Wang, and H. Li, "A two-stage short-term traffic flow prediction method based on AVL and AKNN techniques," *J. Cent. South Univ.*, vol. 22, no. 2, pp. 779–786, 2015.

[56] J. Myung, D.-K. Kim, S.-Y. Kho, and C.-H. Park, "Travel Time Prediction Using *k* Nearest Neighbor Method with Combined Data from Vehicle Detector System and Automatic Toll Collection System," *Transp. Res. Rec. J. Transp. Res. Board*, vol. 2256, pp. 51–59, 2011.

[57] P. Cai, Y. Wang, G. Lu, P. Chen, C. Ding, and J. Sun, "A spatiotemporal correlative k-nearest neighbor model for short-term traffic multistep forecasting," *Transp. Res. Part C Emerg. Technol.*, vol. 62, pp. 21–34, 2016.

[58] J. Yu, G.-L. Chang, H. W. Ho, and Y. Liu, "Variation Based Online Travel Time Prediction Using Clustered Neural Networks," *2008 11th Int. IEEE Conf. Intell. Transp. Syst.*, pp. 85–90, 2008.

[59] D. Park, L. R. Rilett, and G. Han, "Spectral Basis Neural Networks for Real-Time Travel Time Forecasting," *J. Transp. Eng.*, vol. 125, no. 6, pp. 515–523, 1999.

[60] Y. L. Y. Lee, "Freeway travel time forecast using artifical neural networks with cluster method," *2009 12th Int. Conf. Inf. Fusion*, pp. 1331–1338, 2009.

[61] S. Innamaa, "Short-term prediction of travel time using neural networks on an interurban highway," *Transportation (Amst).*, vol. 32, no. 6, pp. 649–669, 2005.

[62] Z. Cui, K. Henrickson, R. Ke, and Y. Wang, "Traffic Graph Convolutional Recurrent Neural Network: A Deep Learning Framework for Network-Scale Traffic Learning and Forecasting," 2018.

[63] Y. Li, R. Yu, C. Shahabi, and Y. Liu, "Diffusion Convolutional Recurrent Neural Network: Data-Driven Traffic Forecasting," pp. 1–16, 2017.

[64] J. Atwood and D. Towsley, "Diffusion-convolutional neural networks," *Adv. Neural Inf. Process. Syst.*, no. Nips, pp. 2001–2009, 2016.

[65] K. Guo, Y. Hu, Z. Qian, Y. Sun, J. Gao, and B. Yin, "Dynamic Graph Convolution Network for Traffic Forecasting Based on Latent Network of Laplace Matrix Estimation," *IEEE Trans. Intell. Transp. Syst.*, vol. 23, no. 2, pp. 1009–1018, 2020.

[66] T. N. Kipf and M. Welling, "Semi-Supervised Classification with Graph Convolutional Networks," pp. 1–14, Sep. 2016.

[67] S. Hochreiter and J. Urgen Schmidhuber, "Long Short-Term Memory," *Neural Comput.*, vol. 9, no. 8, pp. 1735–1780, 1997.

[68] F. Zhuang *et al.*, "A Comprehensive Survey on Transfer Learning," *Proc. IEEE*, vol. 109, no. 1, pp. 43–76, 2021.

[69] "REGIONAL INTEGRATED TRANSPORTATION INFORMATION SYSTEM: A data-driven platform for transportation analysis, monitoring, and data visualization," 2008. [Online]. Available: https://www.ritis.org/traffic/.

[70] F. Pedregosa *et al.*, "Scikit-learn: Machine Learning in Python," *J. Mach. Learn. Res.*, vol. 12, pp. 2825–2830, 2011.

[71] S. van Buuren and K. Groothuis-Oudshoorn, "mice: Multivariate Imputation by Chained Equations in R," *J. Stat. Software, Artic.*, vol. 45, no. 3, pp. 1–67, 2011.

[72] Y. Li, R. Yu, C. Shahabi, and Y. Liu, "Diffusion convolutional recurrent neural network: Data-driven traffic forecasting," *6th Int. Conf. Learn. Represent. ICLR 2018 - Conf. Track Proc.*, pp. 1–16, 2018.

[73] "PyTorch," 2016. [Online]. Available: https://pytorch.org/.